\documentclass[lettersize,journal,10 pt]{IEEEtran}

\usepackage{cite}
\usepackage{amsmath,amssymb,amsfonts}
\usepackage{algorithm}
\usepackage{algpseudocode}

\usepackage{booktabs}

\usepackage{graphicx}
\usepackage{textcomp}
\usepackage{xcolor}
\usepackage{subfigure}
\usepackage{amssymb}
\usepackage{fontenc}
\usepackage{multirow}
\usepackage{url}

\usepackage{soul}
\usepackage{xcolor}

\usepackage[bookmarks=false,hidelinks]{hyperref}

\IEEEoverridecommandlockouts
\begin{document}

	\title{\LARGE \bf Planning-Oriented 3D Scene Completion via Coupled TUDF–Occupancy Representation Learning from Partial Observations
	}

	\author{ Tianyou Yu, Pengfei Zhao, Chao Xu*
	\thanks{
		*Corresponding Author.
		Institute of Cyber-System and Control, College of Control Science and Engineering,
		Zhejiang University, Hangzhou 310027, China.
		E-mail: \texttt{cxu@zju.edu.cn}.
	}
}
	\maketitle
	\thispagestyle{empty}
	\pagestyle{empty}
	

\begin{abstract}
Partial observability remains a fundamental challenge in robotic navigation, where limited sensor coverage and occlusions leave large portions of the environment unobserved. Existing scene completion methods primarily focus on improving incomplete mapping or reconstructing partially observed 3D structures, but rarely investigate how scene completion can be designed to benefit downstream tasks such as path planning. In this work, we propose a path-planning-oriented 3D scene completion framework that moves beyond pure occupancy modeling toward a coupled geometric formulation. Specifically, given partial LiDAR observations as input, the proposed framework jointly predicts completed Truncated Unsigned Distance Field (TUDF)-based continuous geometric representations and voxel-wise occupancy maps. This coupled representation allows the network to better reason about obstacle boundaries and free-space geometry. To fully exploit the synergy between the two representations, we introduce a bidirectionally coupled learning scheme, where TUDF features provide dense geometric guidance to improve occupancy reconstruction, while occupancy features in turn offer complementary structural constraints that refine distance-field estimation. Consequently, the proposed network directly predicts complete occupancy and TUDF representations, allowing seamless integration of TUDF into trajectory planning without post-processing. Extensive experiments on unseen environments demonstrate that the proposed method consistently improves both geometric reconstruction quality and downstream planning performance.
\end{abstract}

\section{Introduction}
	\begin{figure}[t]
	\centering	

	\includegraphics[width=0.5\textwidth]{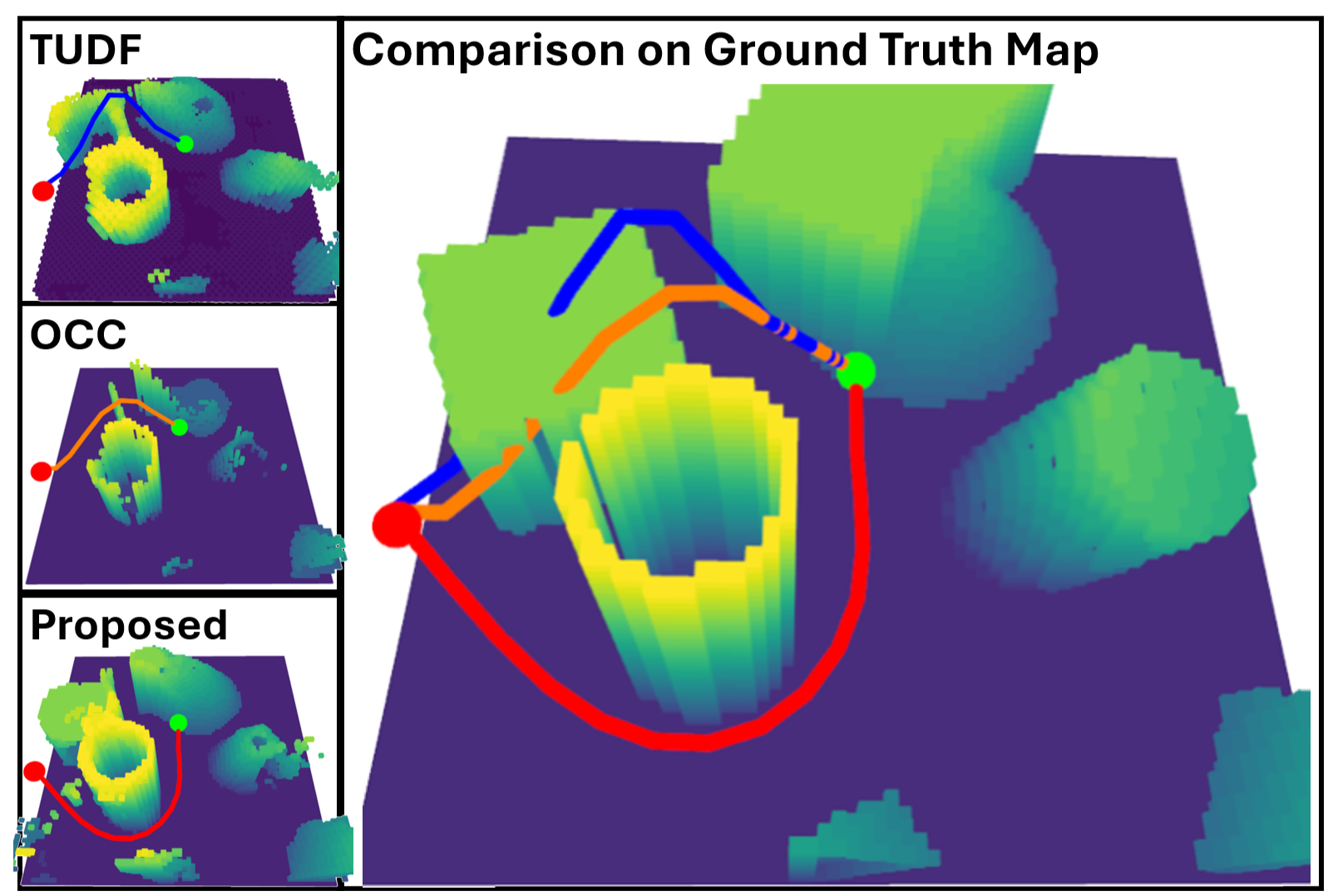}
	\caption{Example path planning results in an unseen environment. From left to right, predicted maps and trajectories generated by TUDF-only (blue), OCC-only (orange), and the proposed method (red) are shown. The trajectories are further evaluated on the ground-truth map, where the green and red spheres indicate the start and goal positions. Occupied voxels are visualized using height-based coloring. In this scenario, TUDF-only and OCC-only fail to recover the complete obstacle geometry, resulting in unsafe trajectories that intersect the true obstacle. In contrast, the proposed method reconstructs a more complete obstacle shape and generates a collision-free trajectory with a larger safety margin.}
	\label{fig:example}
\end{figure} 

Partial observability is a fundamental challenge in autonomous navigation. Limited sensor range and occlusions prevent robots from observing the complete environment, leaving large portions of the scene unknown \cite{nunes2024scaling,georgakis2022uncertainty,reed2024scenesense,wang2021learning}. While many existing methods improve planning and exploration under such uncertainty, they typically operate directly on incomplete observations. An alternative perspective is to infer the missing environmental structure by leveraging prior knowledge and geometric regularities, enabling downstream tasks to reason over a more complete representation of the scene.

Recently, learning-based scene completion and generative models have leveraged geometric priors to infer unobserved structures, enabling more complete reconstruction of objects and scenes \cite{zhu2022nice,chu2023diffcomplete,yu2021pointr}. However, most existing approaches are designed to optimize reconstruction quality, focusing on global shape completion or recovering watertight geometry rather than local map completion from onboard sensor observations. Moreover, they primarily predict occupancy or surface geometry, with little attention paid to constructing reliable continuous distance-field representations that preserve the geometric properties required for path planning. 

To bridge this gap, we propose a planning-oriented scene completion framework that jointly learns occupancy and TUDF representations. Compared with occupancy maps, TUDF provides dense geometric information around obstacle boundaries and can be directly exploited by optimization-based planners. At the same time, unlike signed distance fields (SDF), TUDF avoids the need for reliable inside–outside labels, which are often difficult to obtain from general 3D models or datasets, and enables robust capture of geometric features across diverse objects without being constrained by topology \cite{chu2023diffcomplete}. By jointly learning occupancy and TUDF, our framework combines the complementary strengths of discrete occupancy and continuous geometric representations, producing local maps that are both reconstruction-accurate and planning-friendly.

The main contributions of this work are summarized as follows:

\begin{itemize}
	
	\item We formulate path-planning-oriented 3D scene completion as a coupled TUDF–occupancy representation learning problem, and design a bidirectional interaction architecture that enables effective coupling between continuous distance fields and voxel-wise occupancy representations under partial observations.
	
	\item We demonstrate that the learned TUDF constitutes a continuous, path-planning-grade geometric representation that can be directly used for trajectory planning without post-processing, enabling a streamlined planning pipeline.
	
	\item Extensive experiments on geometric and structured environments show that the proposed method improves both reconstruction accuracy and downstream planning performance under partial observability. The dataset generation pipeline and implementation code will be publicly released upon publication.
	
%
\end{itemize}

\section{Related Work}

Developing planning-oriented scene completion requires understanding both scene representations and how they can be exploited to improve downstream planning performance. Existing works have studied this problem from three perspectives: geometric representations for environment modeling, 3D scene completion for recovering missing structures, and planning-oriented perception for connecting reconstruction with downstream planning.

\subsection{Geometric Scene Representation}

Robotic perception systems rely on geometric representations to model the environment from partial sensor observations. Classical approaches include occupancy grids, TSDF (Truncated Signed Distance Function), which encode either discrete occupancy states or continuous distance-to-surface information \cite{yu2025online}. These representations are typically constructed through Bayesian fusion of multiple observations, such as probabilistic occupancy updates in occupancy grids and weighted averaging schemes in distance field \cite{schmid2023dynablox}. By integrating measurements collected along robot trajectories, they progressively refine a consistent estimate of the environment. However, these methods are inherently conservative, as they primarily rely on observed data and do not explicitly incorporate structural priors.

More recently, neural implicit representations, including neural occupancy fields and neural SDFs, have emerged as a compact alternative for geometric mapping by storing scene information in network parameters \cite{zhu2022nice}. These methods can represent continuous geometry and generalize beyond directly observed measurements. Leveraging learned geometric priors, they can infer plausible structures in partially observed regions and have been successfully integrated into mapping and SLAM systems \cite{wang2023co}. Nevertheless, achieving high-fidelity reconstruction often requires substantial computational resources. Moreover, when trained or optimized primarily from online observations, their ability to recover large missing regions remains limited, and predictions in heavily occluded areas may suffer from hallucinated geometry.

\subsection{3D Scene Completion and Generation}

Unlike traditional approaches that rely solely on observed measurements, learning-based methods leverage data-driven priors to infer complete scene structures from partial observations. Early approaches formulate this problem as deterministic shape completion from depth images or point clouds, learning direct mappings from observed inputs to complete voxel- or point-based representations \cite{yu2021pointr,dai2018scancomplete}.

The introduction of hierarchical U-Net architectures and convolutional neural networks significantly improved the ability to recover missing geometry across multiple spatial resolutions, enabling effective completion of sparse and partially observed maps. More recently, transformer-based architectures have further enhanced global scene understanding by capturing long-range geometric dependencies and structural relationships \cite{yu2021pointr}. Several works have also demonstrated that incorporating semantic cues or label-conditioned priors can improve completion quality in both indoor and outdoor environments by providing object-level contextual information. Furthermore, generative models, particularly diffusion-based approaches \cite{reed2024scenesense,martyniuk2025lidpm,luo2021diffusion}, have shown strong capabilities in modeling uncertainty and generating plausible structures in unobserved regions that remain consistent with available observations.

Despite these advances, most existing methods focus on global shape consistency and are evaluated primarily using reconstruction metrics. They rarely consider the requirements of robotic navigation, where accurate local geometry, free-space estimation, and obstacle boundary fidelity are critical for downstream planning under partial observability.

\subsection{Planning-Oriented Perception}

Path planning depends critically on the quality of environmental perception. While end-to-end approaches directly predict trajectories from sensor observations, perception–planning frameworks remain prevalent in robotics due to their interpretability and modularity. As a result, many learning-based navigation systems \cite{yang2025driving,zheng2024occworld} leverage geometric representations as intermediate supervision to improve planning performance.

Without leveraging structural priors, planning under partial observability faces limitations similar to those of classical mapping approaches, where unobserved regions are typically treated as either occupied or free space \cite{tordesillas2021faster}. While such assumptions simplify planning, they often lead to overly conservative behavior or increased collision risk.

Learning-based perception methods alleviate this issue by incorporating data-driven priors, enabling predictions beyond directly observed measurements. To further account for uncertainty, several works employ ensemble models or probabilistic predictors to estimate epistemic uncertainty and guide planners away from uncertain regions \cite{georgakis2022uncertainty}. However, such uncertainty estimates primarily reflect prediction disagreement induced by dataset bias and limited model capacity, rather than explicitly reasoning about the underlying geometric structure of the environment.

Distance-field representations, such as SDFs, are particularly attractive for trajectory optimization because they provide continuous geometric information in the form of signed distances \cite{jacquet2025neural}. Nevertheless, most learning-based completion methods focus on occupancy prediction and do not directly reconstruct a planning-oriented distance field \cite{wang2021learning,dhami2023pred,reed2025online}. This is partly because retrieving the sign information from arbitrary 3D CAD models is non-trivial. By using TUDF, we can robustly capture the geometric features of different objects without being constrained by topology \cite{chu2023diffcomplete,zhang2023surface}.

Above observations motivate the development of scene representations that simultaneously model occupancy and continuous geometric structure, allowing the reconstructed map to be directly utilized for trajectory planning while maintaining accurate obstacle and free-space estimation under partial observability.

\section{Method}
\label{sec:method}
\subsection{Training Data Generation}
\label{subsec:data_gen}
The proposed network takes a partial LiDAR observation as input and predicts the 
corresponding complete occupancy map and TUDF representation. Accordingly, each 
training sample is formulated as $\mathcal{S}_i=(X_i,O_i,D_i)$, where $X_i$ denotes the voxelized partial LiDAR observation, $O_i$ represents the complete occupancy ground truth, and $D_i$ denotes the corresponding TUDF representation. All three representations are defined in the same voxel coordinate system with identical spatial resolution, enabling voxel-wise supervision between the input observation and the predicted outputs.

Specifically, the input observation is represented as a binary voxel grid $X_i \in \{0,1\}^{H\times W\times D}$. The complete occupancy map $O_i$ is obtained from the underlying environment, and the TUDF ground truth is generated from the occupancy representation using a Euclidean distance transform. For each voxel location $v$, the unsigned distance is computed as
\begin{equation}
D_i(v)=
\min_{o\in \mathcal{O}_i}
\|v-o\|_2 ,
\end{equation}
where $\mathcal{O}_i$ denotes the set of occupied voxels. The distance values are 
then truncated within a predefined range $d_{\max}$:
\begin{equation}
D_i(v)=\min(D_i(v),d_{\max}).
\end{equation}

To collect training samples, synthetic environments are constructed in Unreal 
Engine. Various geometric obstacles, including walls, pillars, and cluttered 
structures, are randomly placed to create diverse navigation scenarios, as shown in 
Figure~\ref{fig:sim_envs}. For each environment, an RRT planner generates 
collision-free trajectories between randomly sampled start and goal positions. LiDAR 
measurements are simulated along these trajectories to obtain partial observations 
from different viewpoints. Scenes with insufficient geometric complexity are removed 
to ensure sufficient obstacle diversity and structural variation.

\subsection{Coupled TUDF–Occupancy Learning}

\begin{figure}[t]
	\centering
	\includegraphics[width=0.5\textwidth]{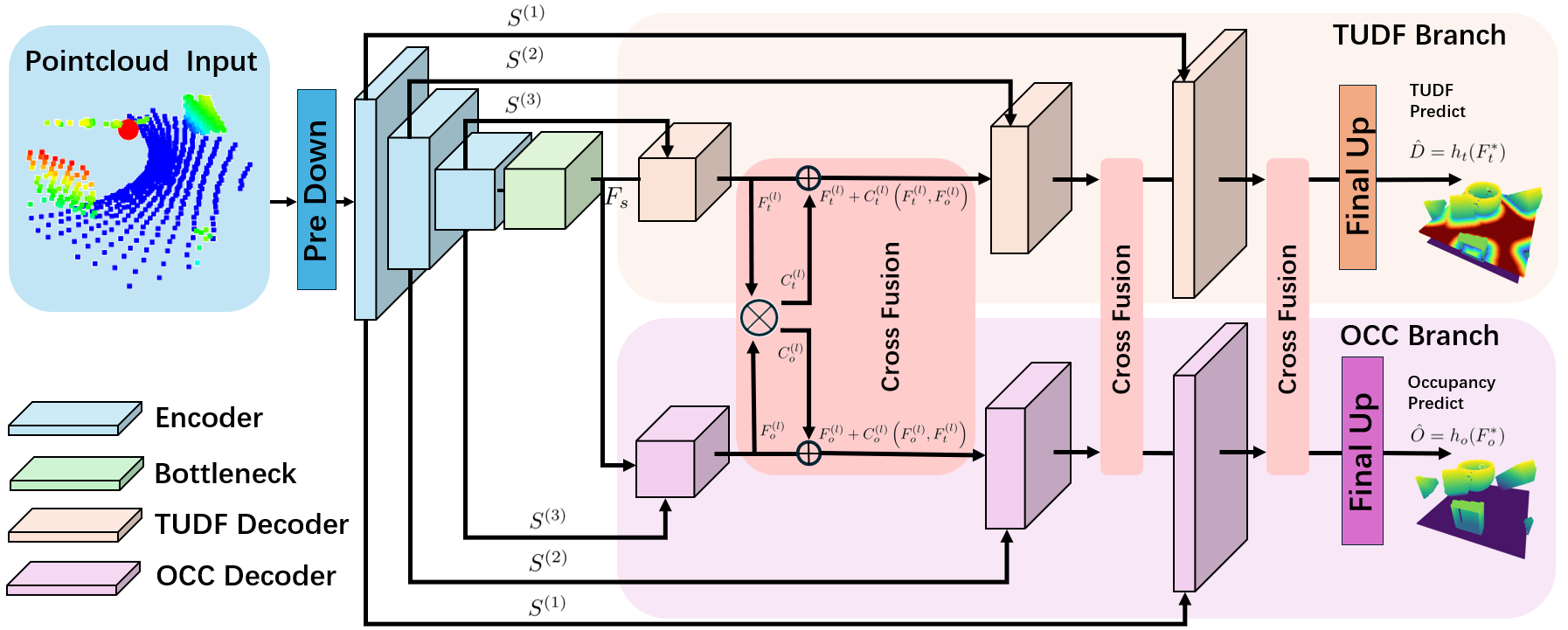}
	\caption{Overview of the proposed dual-decoder reconstruction network with bidirectional CrossFusion. A shared encoder extracts hierarchical geometric features, which are progressively decoded by TUDF and occupancy branches with bidirectional CrossFusion at multiple decoder stages.}
	\label{fig:network_framework}
\end{figure}

Given a partial point cloud, we first voxelize it into a binary occupancy grid
$X\in\mathbb{R}^{H\times W\times D}$.
A shallow pre-processing module extracts initial geometric features,
which are then encoded by a shared 3D encoder to capture hierarchical scene
representations. Specifically, the encoder progressively reduces the spatial resolution through multi-scale feature extraction and employs skip connections to preserve fine-grained geometric details, producing multi-scale skip features
$\{S^{(1)},S^{(2)},S^{(3)}\}$. A self-attention module is further introduced
at the bottleneck to capture long-range spatial dependencies, producing a
shared latent representation $F_s$.

Instead of predicting different representations from a single decoder, we design a 
\emph{hierarchically coupled dual-decoder} architecture to jointly recover continuous 
geometry and discrete occupancy. The shared latent representation is progressively 
decoded through coarse-to-fine stages with encoder skip connections. Specifically, the 
first decoding stage initializes the two branches:
\begin{equation}
F_t^{(3)} = f_t^{(3)}(F_s,S^{(3)}), \qquad
F_o^{(3)} = f_o^{(3)}(F_s,S^{(3)}),
\end{equation}
followed by hierarchical feature refinement:

\begin{equation}
F_t^{(l)}=f_t^{(l)}(F_t^{(l+1)},S^{(l)}),
F_o^{(l)}=f_o^{(l)}(F_o^{(l+1)},S^{(l)}),
l=2,1,
\end{equation}

To enable bidirectional interaction throughout the decoding process, CrossFusion is 
applied at each decoder stage after feature refinement and before forwarding the 
features to the next decoding stage:
\begin{equation}
\begin{aligned}
F_t^{(l)} &\leftarrow 
F_t^{(l)}
+
C_t^{(l)}
\left(
F_t^{(l)},F_o^{(l)}
\right),\\
F_o^{(l)} &\leftarrow 
F_o^{(l)}
+
C_o^{(l)}
\left(
F_o^{(l)},F_t^{(l)}
\right).
\end{aligned}
\end{equation}
where $C_t^{(l)}(\cdot)$ and $C_o^{(l)}(\cdot)$ denote the learnable 
bidirectional CrossFusion operators at decoder stage $l$. Specifically, each CrossFusion module concatenates the features from the two branches 
along the channel dimension and applies lightweight point-wise ($1\times1\times1$) 3D convolutions with nonlinear activation to learn adaptive feature transformations. This design allows the network to automatically determine which geometric or structural features should be exchanged between the TUDF and occupancy branches.

Through this progressive coarse-to-fine decoding strategy, geometric distance estimation and occupancy reasoning continuously exchange complementary information at multiple spatial resolutions. Finally, the refined features at the last decoder stage are converted into TUDF and occupancy predictions:
\begin{equation}
\hat{D} = h_t\!\left(F_t^{(1)}\right), \qquad
\hat{O} = h_o\!\left(F_o^{(1)}\right).
\end{equation}
The $h_t$ and $h_o$ correspond to the final prediction heads implemented as the Final-Up modules. Here, $\hat{D} \in \mathbb{R}^{H \times W \times D}$ denotes the predicted TUDF, and $\hat{O} \in [0,1]^{H \times W \times D}$ represents the voxel-wise occupancy probability map defined over the same 3D grid.

The proposed architecture directly predicts complete TUDF and occupancy representations from partial LiDAR observations. Through bidirectional feature interaction, TUDF and occupancy representations are jointly optimized, where geometric cues enhance occupancy boundary reconstruction and occupancy constraints refine distance-field estimation. This coupled learning strategy improves geometric reconstruction and downstream path planning performance. The learned TUDF provides a continuous geometric representation that can be directly integrated into trajectory planning without post-processing. 


\subsection{Learning Objective}

The proposed network is trained in an end-to-end manner by jointly optimizing the occupancy prediction and TUDF regression tasks. The overall objective is defined as

\begin{equation}
\mathcal{L}
=
\lambda_{occ}\mathcal{L}_{occ}
+
\lambda_{tudf}\mathcal{L}_{tudf},
\end{equation}

where $\lambda_{occ}$ and $\lambda_{tudf}$ balance the contributions of the two tasks. Here, $\hat{O}$ and $O$ denote the predicted and ground-truth occupancy fields, respectively, and $\hat{D}$ and $D$ denote the predicted and ground-truth TUDF.

The occupancy branch is formulated as a voxel-wise binary classification problem. Since free-space voxels significantly outnumber occupied voxels, a weighted binary cross-entropy loss is adopted,

\begin{equation}
\mathcal{L}_{occ}
=
-\frac{1}{N}
\sum_i
w^{o}_i
\left[
o_i\log(\hat{o}_i)
+
(1-o_i)\log(1-\hat{o}_i)
\right],
\end{equation}
where $o_i \in O$ and $\hat{o}_i \in \hat{O}$ denote the ground-truth and predicted occupancy probabilities at voxel $i$, respectively. $N$ denotes the number of voxels, and $w^{o}_i$ assigns a larger weight to occupied voxels to alleviate class imbalance. 

The TUDF branch is optimized using a weighted Smooth-L1 regression loss,

\begin{equation}
\mathcal{L}_{reg}
=
\frac{1}{N}
\sum_i
w^{t}_i
\,
\mathrm{SmoothL1}
(\hat d_i,d_i),
\end{equation}
where $d_i \in D$ and $\hat{d}_i \in \hat{D}$ denote the ground-truth and predicted TUDF values at voxel $i$, respectively, and $w_i^{t}$ assigns larger weights to voxels closer to obstacle surfaces.

To further encourage the predicted distance field to satisfy the geometric property of a true distance field, an Eikonal regularization term is introduced within the truncation region:
\begin{equation}
\mathcal{L}_{eik}
=
\frac{1}{\sum_i m_i}
\sum_i
m_i
\left|
\|\nabla\hat d_i\|_2-1
\right|,
\end{equation}
where $m_i$ is a binary mask indicating whether the voxel distance is within the 
truncation range, and $\nabla\hat d_i$ is computed using finite differences along the 
three voxel axes.

The final TUDF objective is therefore
\begin{equation}
\mathcal{L}_{tudf}
=
\mathcal{L}_{reg}
+
\lambda_{eik}\mathcal{L}_{eik},
\end{equation}
where $\lambda_{eik}$ controls the contribution of the Eikonal regularization. The regression loss encourages accurate distance estimation, while the Eikonal constraint promotes locally consistent gradients and smoother geometric structures, resulting in more reliable distance fields for downstream trajectory planning.
\subsection{Planning Evaluation}

To evaluate the reconstructed maps for downstream planning, we perform trajectory 
planning using the predicted occupancy map and TUDF representation. The occupancy 
prediction is first converted into a binary collision map, based on which a 3D A* 
planner generates an initial collision-free path
$Q=\{\mathbf{p}_0,\mathbf{p}_1,\cdots,\mathbf{p}_T\}$,
where $\mathbf{p}_0$ and $\mathbf{p}_T$ are the given start and goal positions. 
The generated path is then refined by trajectory optimization using the predicted 
TUDF. Specifically, the optimized trajectory is obtained by
\begin{equation}
Q^*=\arg\min_Q
J_{\mathrm{smooth}}
+\lambda_c J_{\mathrm{clear}},
\end{equation}
where the optimization is performed only over intermediate waypoints while keeping 
the start and goal fixed. The smoothness term encourages locally smooth trajectories,
while the clearance term penalizes trajectories approaching obstacles:
\begin{equation}
J_{\mathrm{smooth}}
=
\sum_i
\|
\mathbf{p}_{i+1}-2\mathbf{p}_i+\mathbf{p}_{i-1}
\|^2,
\end{equation}
\begin{equation}
J_{\mathrm{clear}}
=
\sum_i
\max(0,d_s-\hat D(\mathbf p_i))^2,
\end{equation}
where $\hat D(\mathbf p_i)$ denotes the predicted TUDF value at waypoint 
$\mathbf p_i$, and $d_s$ is the desired safety clearance. The clearance cost is 
activated only when the predicted distance falls below the safety threshold, thereby 
encouraging collision-free trajectories with sufficient obstacle margins. Planning 
performance is evaluated using path length, minimum clearance, average clearance, and 
success rate.

\section{Experiments}

\begin{figure}[t]
	\centering
	\subfigure[random geometric]{\includegraphics[width=1.7in]{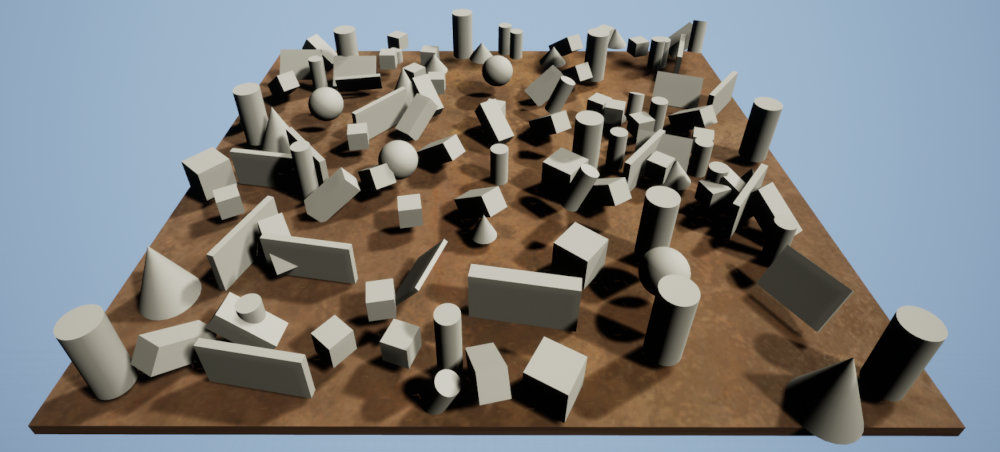}}
	\subfigure[dense pillar]{\includegraphics[width=1.5in]{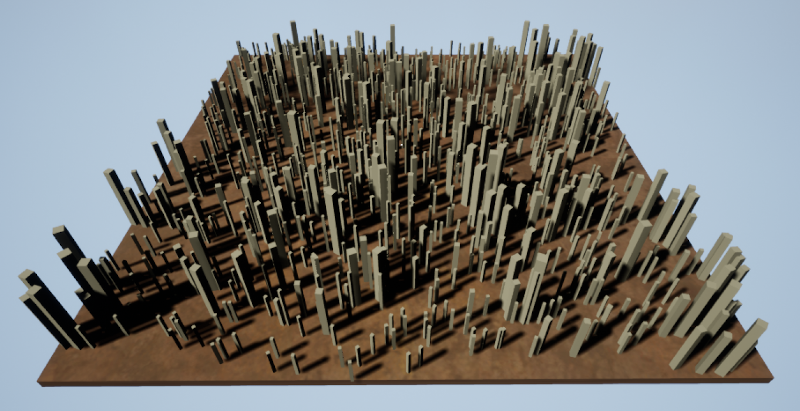}}
	\caption{Simulated scences for generating sensor measurements and ground truth map.}
	\label{fig:sim_envs}
\end{figure}

In this section, we evaluate the proposed framework in two representative types of environments: random geometric environments (Figure~\ref{fig:sim_envs}(a)) and dense pillar environments (Figure~\ref{fig:sim_envs}(b)). The former contains random obstacle geometries and layouts to evaluate the generalization ability of scene reconstruction, while the latter introduces densely distributed obstacles and narrow passages. Specifically, we seek to answer the following questions:

\begin{itemize}
	\item \textbf{Q1}: Does jointly learning occupancy and TUDF improve reconstruction performance compared with single-task baselines?
	\item \textbf{Q2}: Can the improved reconstruction quality benefit downstream motion planning in unknown environments?
	\item \textbf{Q3}: How does the quality of the reconstructed geometric representation affect downstream trajectory planning?
\end{itemize}

\textbf{Q1} is addressed through qualitative and quantitative reconstruction comparisons with single-task baselines in Sections~\ref{subsec:visual_recon_results} and~\ref{subsec:quan_recon_results}. \textbf{Q2} is evaluated through downstream planning experiments in partially observed environments in Section~\ref{subsec:planning_results}. Finally, \textbf{Q3} examines how reconstruction quality translates into downstream trajectory performance by jointly analyzing the reconstruction and planning results.

\subsection{Implementation Details}
%
The dataset is collected in the simulation environments introduced in Section~\ref{subsec:data_gen}, with 2500 training samples per scene category and 500 samples from previously unseen environments for testing. All reconstruction inference and downstream planning evaluations are performed exclusively on the test set. All methods use a voxel resolution of $0.2\,\mathrm{m}$ and an $80\times80\times40$ voxel local map, corresponding to $16\,\mathrm{m}\times16\,\mathrm{m}\times8\,\mathrm{m}$. The simulated LiDAR has a $10\,\mathrm{m}$ sensing range, 120 horizontal and 16 vertical scanning lines, and a $60^\circ$ vertical field of view, with a $30^\circ$ downward pitch relative to the robot body frame. Measurements outside the local map volume are excluded during voxelization.

We compare three map prediction strategies: \textbf{OCC-only}, a single-branch 3D U-Net trained solely for occupancy prediction \cite{wang2021learning}; \textbf{TUDF-only}, using the same architecture for TUDF regression; and \textbf{Proposed}, which jointly predicts occupancy and TUDF through coupled prediction branches. All models use the same 3D U-Net backbone with a base channel width of 8, three encoder-decoder levels, and self-attention at the bottleneck. All models are trained under identical optimization settings on an NVIDIA RTX 4070 GPU, with early stopping after 10 consecutive epochs without validation improvement and the best validation checkpoint selected for testing. Given an input volume of $80\times80\times40$, the Proposed model achieves approximately $5.5 \mathrm{ms}$ inference time per forward pass.

To ensure a fair and consistent evaluation across different representations, all methods are assessed using both occupancy-based and distance-field-based metrics. For the OCC-only model, voxels with predicted occupancy probabilities greater than 0.85 are considered occupied. Its corresponding TUDF is obtained by applying Euclidean distance transform to the predicted occupied voxels, the resulting distance field is treated as a derived geometric proxy for consistent evaluation and visualization across representations. Similarly, For the TUDF-only model, occupancy labels are derived by thresholding the predicted distance field, where voxels with predicted distances smaller than $0.2\,\mathrm{m}$ are classified as occupied. The proposed framework directly predicts both occupancy probabilities and TUDF values in a single forward pass, and occupancy is obtained using the same probability threshold of $0.85$. For all methods, the converted occupancy maps and TUDF fields are further used for metric computation and qualitative visualization in a unified voxel grid space, ensuring consistent evaluation across different representations.


The TUDF truncation distance is set to $d_{\max}=2.0\,\mathrm{m}$, with loss weights $\lambda_{occ}=0.1$, $\lambda_{tudf}=1.0$, and $\lambda_{eik}=0.05$. Occupied and free-space voxels are weighted by $w_i^{o}=5$ and $1$, respectively, while TUDF voxels within $0.5\,\mathrm{m}$ of obstacle surfaces use $w_i^{t}=2$ and others use $w_i^{t}=1$. For planning, $\lambda_c=1.0$, the A* step size is $0.6\,\mathrm{m}$ (3$\times$ the voxel resolution), and trajectory optimization runs for 50 iterations.
\subsection{Visual Reconstruction Results}
\label{subsec:visual_recon_results}
	
\begin{figure*}[t]
	\centering
	\includegraphics[width=0.85\textwidth]{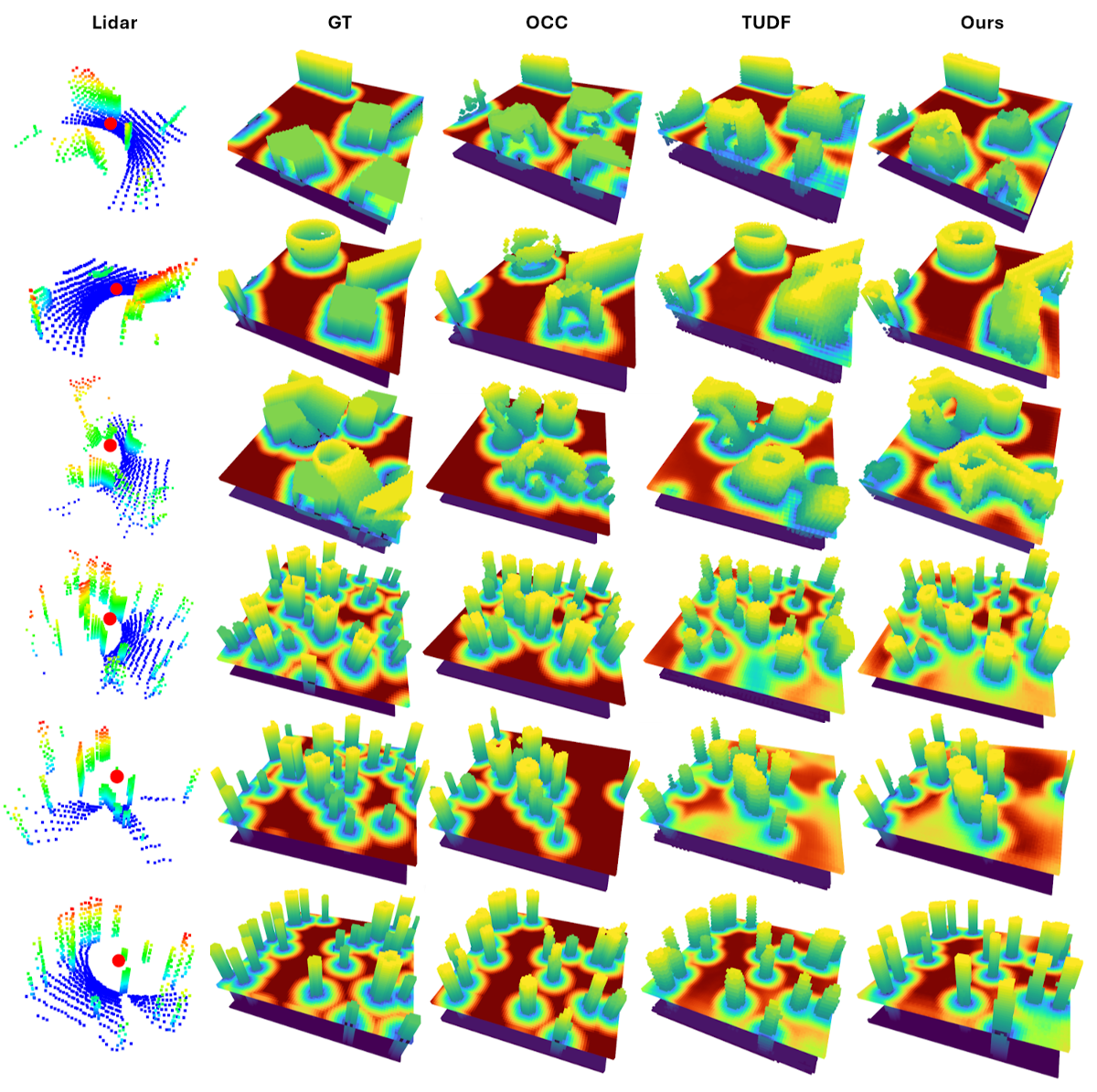}
	\caption{Qualitative comparison of reconstruction results in unseen environments. The first three rows show geometric scenes, while the last three rows show dense pillar scenes. From left to right: input LiDAR observation, ground-truth map, OCC-only prediction, TUDF-only prediction, and the proposed joint TUDF-OCC prediction. The red sphere denotes the LiDAR position. Input points are colored by height, and occupancy maps are visualized using Turbo colormap. For TUDF visualization, only voxels within a selected height range are displayed to reduce occlusion, with colors indicating distances from $0\,\mathrm{m}$ (blue) to truncated distance $d_{max}=2\,\mathrm{m}$ (red).}
	\label{fig:compare}
\end{figure*}

Figure~\ref{fig:compare} presents representative reconstruction results in previously unseen environments, including both geometric scenes and pillar environments. In geometric environments, where objects such as spheres, boxes, and cylinders exhibit regular structures, the OCC-only approach is able to recover the overall scene layout. However, it often produces incomplete or fragmented reconstructions when only partial observations are available, leading to inaccurate obstacle boundaries and missing geometric details. In contrast, the TUDF-only method benefits from continuous distance-field supervision, producing smoother and more complete geometric reconstruction. However, under sparse observations it tends to over-complete nearby regions, resulting in spurious occupancy in free space.


In pillar environments, where obstacles exhibit relatively simple local geometry, the network can effectively learn their structural patterns given sufficient observations. However, for partially observed or unobserved obstacles, limited geometric evidence makes their spatial extent difficult to estimate accurately. Consequently, OCC-only tends to preserve observed obstacle regions and partially complete visible structures, but often fails to recover missing obstacles under sparse observations, resulting in large regions predicted as free space. The TUDF converted from these occupancy predictions therefore tends to assign maximum truncated distances to such regions, since no nearby obstacle boundaries are explicitly indicated. In contrast, TUDF-only directly optimizes the continuous distance representation and, although it may not explicitly reconstruct complete obstacle geometry in unobserved regions, tends to predict lower distances around potentially occupied areas. While this can reduce the accuracy of the recovered obstacle shape, it provides more conservative geometric cues in uncertain regions, which can benefit downstream planning.

Across both geometric and pillar environments, the proposed framework consistently combines the complementary strengths of the two representations. The OCC branch preserves reliable obstacle structures, while the TUDF branch provides continuous geometric guidance in sparsely observed regions. Their bidirectional interaction produces more complete reconstructions while mitigating both the conservative predictions of the OCC-only model and the over-expansion tendency of the TUDF-only model.

For completely unobserved areas, all methods, tend to predict free space due to the absence of supporting observations. This highlights the fundamental limitation of deterministic reconstruction under extreme sparsity.

\subsection{Quantitative Reconstruction Results}
\label{subsec:quan_recon_results}
\begin{table}[t]
	\centering
	\caption{Performance comparison on geometric and pillar scenes. For Distance Bias, values closer to zero indicate lower systematic errors in distance estimation.}
	\label{tab:ablation_and_pillar}
	\renewcommand{\arraystretch}{1.1}
	\footnotesize
	
	\resizebox{\columnwidth}{!}{
		\begin{tabular}{llccc}
			\hline
			\textbf{Scene} & \textbf{Metric} & \textbf{TUDF} & \textbf{OCC} & \textbf{Proposed} \\
			\hline
			
			\multirow{10}{*}{Geometric}
			
			& RMSE $\downarrow$
			& \textbf{0.317 $\pm$ 0.097}
			& 0.437 $\pm$ 0.123
			& 0.321 $\pm$ 0.097 \\
			
			& MAE $\downarrow$
			& \textbf{0.162 $\pm$ 0.068}
			& 0.203 $\pm$ 0.081
			& \textbf{0.162 $\pm$ 0.064} \\
			
			& Near-RMSE $\downarrow$
			& 0.310 $\pm$ 0.111
			& 0.576 $\pm$ 0.156
			& \textbf{0.282 $\pm$ 0.113} \\
			
			& Near-MAE $\downarrow$
			& 0.157 $\pm$ 0.064
			& 0.313 $\pm$ 0.114
			& \textbf{0.141 $\pm$ 0.062} \\
			
			& Distance Bias (m)
			& \textbf{0.004 $\pm$ 0.050}
			& 0.188 $\pm$ 0.082
			& -0.028 $\pm$ 0.047 \\
			
			& Near Distance Bias (m)
			& 0.101 $\pm$ 0.064
			& 0.305 $\pm$ 0.115
			& \textbf{0.064 $\pm$ 0.065} \\
			
			& IoU $\uparrow$
			& 0.526 $\pm$ 0.068
			& 0.575 $\pm$ 0.081
			& \textbf{0.601 $\pm$ 0.080} \\
			
			& F1 Score $\uparrow$
			& 0.687 $\pm$ 0.060
			& 0.727 $\pm$ 0.067
			& \textbf{0.747 $\pm$ 0.064} \\
			
			& Precision $\uparrow$
			& 0.830 $\pm$ 0.046
			& \textbf{0.966 $\pm$ 0.029}
			& 0.867 $\pm$ 0.043 \\
			
			& Recall $\uparrow$
			& 0.590 $\pm$ 0.077
			& 0.587 $\pm$ 0.083
			& \textbf{0.662 $\pm$ 0.090} \\
			
			\hline
			
			\multirow{10}{*}{Pillar}
			
			& RMSE $\downarrow$
			& 0.355 $\pm$ 0.069
			& 0.485 $\pm$ 0.095
			& \textbf{0.329 $\pm$ 0.068} \\
			
			& MAE $\downarrow$
			& 0.214 $\pm$ 0.050
			& 0.251 $\pm$ 0.072
			& \textbf{0.191 $\pm$ 0.050} \\
			
			& Near-RMSE $\downarrow$
			& 0.344 $\pm$ 0.075
			& 0.574 $\pm$ 0.112
			& \textbf{0.338 $\pm$ 0.079} \\
			
			& Near-MAE $\downarrow$
			& 0.186 $\pm$ 0.047
			& 0.308 $\pm$ 0.087
			& \textbf{0.177 $\pm$ 0.049} \\
			
			& Distance Bias (m)
			& \textbf{0.016 $\pm$ 0.061}
			& 0.238 $\pm$ 0.077
			& 0.026 $\pm$ 0.060 \\
			
			& Near Distance Bias (m)
			& 0.134 $\pm$ 0.053
			& 0.304 $\pm$ 0.087
			& \textbf{0.128 $\pm$ 0.054} \\
			
			& IoU $\uparrow$
			& 0.440 $\pm$ 0.043
			& 0.556 $\pm$ 0.058
			& \textbf{0.559 $\pm$ 0.058} \\
			
			& F1 Score $\uparrow$
			& 0.610 $\pm$ 0.042
			& 0.713 $\pm$ 0.048
			& \textbf{0.715 $\pm$ 0.049} \\
			
			& Precision $\uparrow$
			& 0.681 $\pm$ 0.037
			& \textbf{0.976 $\pm$ 0.035}
			& 0.890 $\pm$ 0.039 \\
			
			& Recall $\uparrow$
			& 0.557 $\pm$ 0.065
			& 0.564 $\pm$ 0.057
			& \textbf{0.601 $\pm$ 0.065} \\
			
			\hline
		\end{tabular}
	}
\end{table}

The reconstruction quality is evaluated using both occupancy-based and distance-field-based metrics. For occupancy prediction, we report the Intersection-over-Union (IoU), Precision, Recall, and F1 score after thresholding the predicted occupancy probabilities. For TUDF prediction, we report the Root Mean Square Error (RMSE), Mean Absolute Error (MAE), and the signed Distance Bias. Since accurate geometry near obstacle surfaces is particularly important for trajectory planning, we further report the corresponding Near-RMSE, Near-MAE, and Near Distance Bias, which are computed over voxels within $1.0\,\mathrm{m}$ of obstacle surfaces. The quantitative results are summarized in Table~\ref{tab:ablation_and_pillar}.

The distance-field metrics reveal that reconstructing a TUDF solely from predicted occupancy voxels is insufficient for accurate geometric estimation. Although the OCC-only model achieves the highest precision and a relatively high IoU, its reconstructed TUDF exhibits substantially larger RMSE, Near-RMSE, and distance bias. This is because occupancy prediction is inherently conservative: partially observed obstacles are often incompletely reconstructed, and the resulting missing occupied voxels have only a limited impact on occupancy metrics but significantly alter the distance field after conversion. Consequently, the reconstructed TUDF systematically overestimates the distance to nearby obstacles, leading to the large positive bias observed in Table~\ref{tab:ablation_and_pillar}. Such systematic overestimation is particularly undesirable for trajectory planning, as regions that are actually close to obstacles may be incorrectly considered safe, increasing the risk of collision.

In contrast, both the TUDF-only model and the proposed framework produce substantially lower distance bias owing to the continuous nature of TUDF supervision. Rather than assigning the truncation distance to unobserved regions, the network is able to infer plausible distance values based on surrounding geometric context, as illustrated in Fig.~\ref{fig:compare}. This results in more accurate distance estimation, especially in partially observed areas.

Finally, the proposed framework achieves the highest IoU and F1 score while simultaneously obtaining the lowest Near-RMSE and Near-MAE. These results indicate that TUDF learning improves occupancy prediction by providing dense geometric cues, whereas occupancy supervision further refines distance estimation around obstacle boundaries. The bidirectional interaction between the two representations therefore leads to more accurate geometric reconstruction, particularly in near-surface regions where precise distance estimation is critical for downstream trajectory planning.

\subsection{Planning with Predicted Distance Fields}
\label{subsec:planning_results}
\label{experiments:planning}
	\begin{figure*}[t]
		\centering
		
		\includegraphics[width=0.875\textwidth]{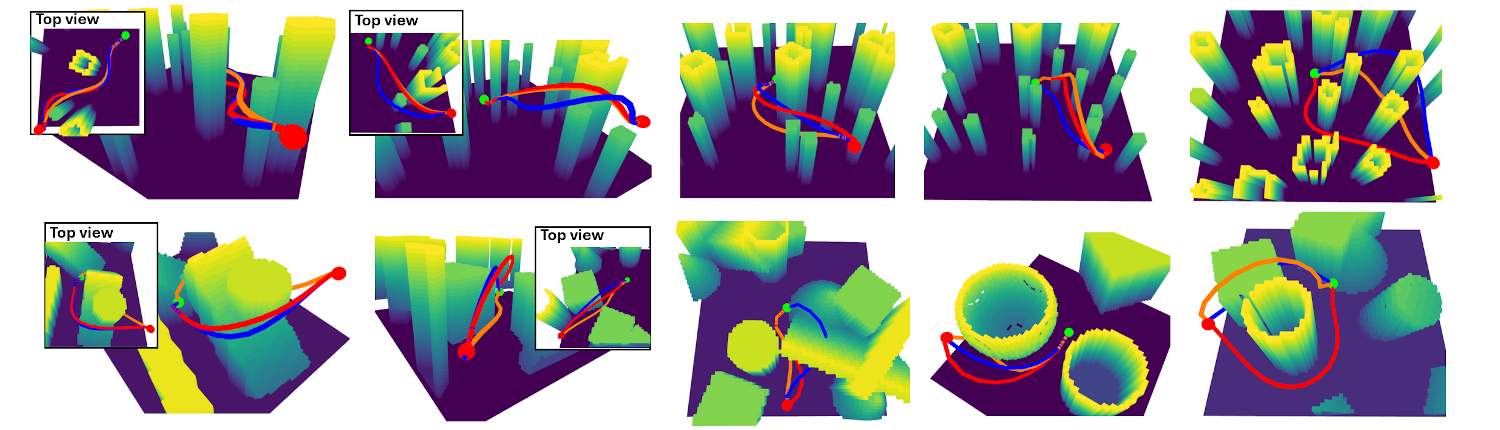}
		\caption{Comparison of path planning results in unseen environments. For brevity, only planned trajectories are visualized on the ground-truth maps instead of the predicted maps. The top row shows representative examples from the pillar environment, while the bottom row corresponds to the random geometric environment. Trajectories generated by the TUDF-only, OCC-only, and proposed methods are shown in blue, orange, and red, respectively. Occupied voxels are colored using the Viridis colormap according to their height along the $z$-axis. The green and red spheres indicate the start and goal positions.}
		\label{fig:planning_compare}
	\end{figure*}


We further evaluate the three reconstruction methods in a path planning benchmark. For each test scene, the start position is fixed at the map center, corresponding to the simulated sensor location, while goal positions are randomly sampled near the map boundary to require navigation beyond the observed region. Planning is performed in a single shot without replanning, allowing us to systematically evaluate the effect of reconstruction quality on trajectory generation. The evaluation traverses the test samples until 2000 valid planning trials are collected for each scene category. Cases where no feasible solution is found using the ground-truth map are excluded.

The resulting occupancy map and TUDF are used by the planning module described in Section~\ref{sec:method}.D to generate collision-free trajectories. All planned trajectories are subsequently evaluated in the corresponding ground-truth environment.

\begin{table}[t]
	\centering
	\caption{Path planning performance in geometric and pillar environments. Best results are highlighted in bold.}
	\label{tab:planning_all}
	\renewcommand{\arraystretch}{1.1}
	\footnotesize
	
	\resizebox{\columnwidth}{!}{
		\begin{tabular}{llccc}
			\hline
			\textbf{Scene} & \textbf{Metric} & \textbf{TUDF} & \textbf{OCC} & \textbf{Proposed} \\
			\hline
			
			\multirow{4}{*}{Geometric}
			
			& Path Length $\downarrow$
			& 9.377$\pm$1.279
			& \textbf{9.156$\pm$1.020}
			& 9.404$\pm$1.281 \\
			
			& Min Clearance $\uparrow$
			& 0.985$\pm$0.713
			& 0.917$\pm$0.737
			& \textbf{1.007$\pm$0.698} \\
			
			& Avg Clearance $\uparrow$
			& \textbf{1.376$\pm$0.675}
			& 1.347$\pm$0.657
			& 1.371$\pm$0.683 \\
			
			& Success Rate $\uparrow$
			& 78.7\%
			& 77.8\%
			& \textbf{80.6\%} \\
			
			\hline
			
			\multirow{4}{*}{Pillar}
			
			& Path Length $\downarrow$
			& 8.639$\pm$3.885
			& \textbf{7.873$\pm$1.572}
			& 8.525$\pm$3.881 \\
			
			& Min Clearance $\uparrow$
			& 0.750$\pm$0.403
			& 0.725$\pm$0.418
			& \textbf{0.816$\pm$0.370} \\
			
			& Avg Clearance $\uparrow$
			& 1.312$\pm$0.356
			& 1.310$\pm$0.268
			& \textbf{1.326$\pm$0.352} \\
			
			& Success Rate $\uparrow$
			& 84.6\%
			& 76.9\%
			& \textbf{88.5\%} \\
			
			\hline
		\end{tabular}
	}
\end{table}


To evaluate downstream planning performance, we report path length, clearance and success rate. Clearance is computed as the Euclidean distance to the nearest obstacle using the ground-truth TUDF, with minimum and average clearance taken along each trajectory. Quantitative results are summarized in Table~\ref{tab:planning_all}, with representative examples shown in Fig.~\ref{fig:planning_compare}.

The OCC-only method produces the shortest trajectories. However, this is primarily caused by its positive distance bias, which systematically overestimates obstacle distances during trajectory optimization. As a result, the planner tends to generate trajectories that pass closer to obstacles. Combined with the incomplete reconstruction of partially observed obstacles, this leads to reduced obstacle clearance and more collision failures, despite the relatively high occupancy accuracy.

The TUDF-only method benefits from more accurate distance estimation, allowing the planner to maintain larger obstacle clearances. However, the occupancy map recovered from the predicted TUDF is occasionally incomplete. Consequently, the initial trajectory may pass through missing obstacles, and because the TUDF is unsigned, the optimizer cannot effectively recover from such invalid initializations.

In contrast, the proposed method jointly learns occupancy and TUDF representations, leading to both more complete obstacle reconstruction and more accurate distance estimation near obstacle boundaries. As a result, the planner maintains sufficient safety margins without becoming overly conservative, achieving the highest success rate while providing a better balance between planning efficiency and safety.

\begin{figure}[t]
	\centering
	\subfigure[]{\includegraphics[width=1.1in]{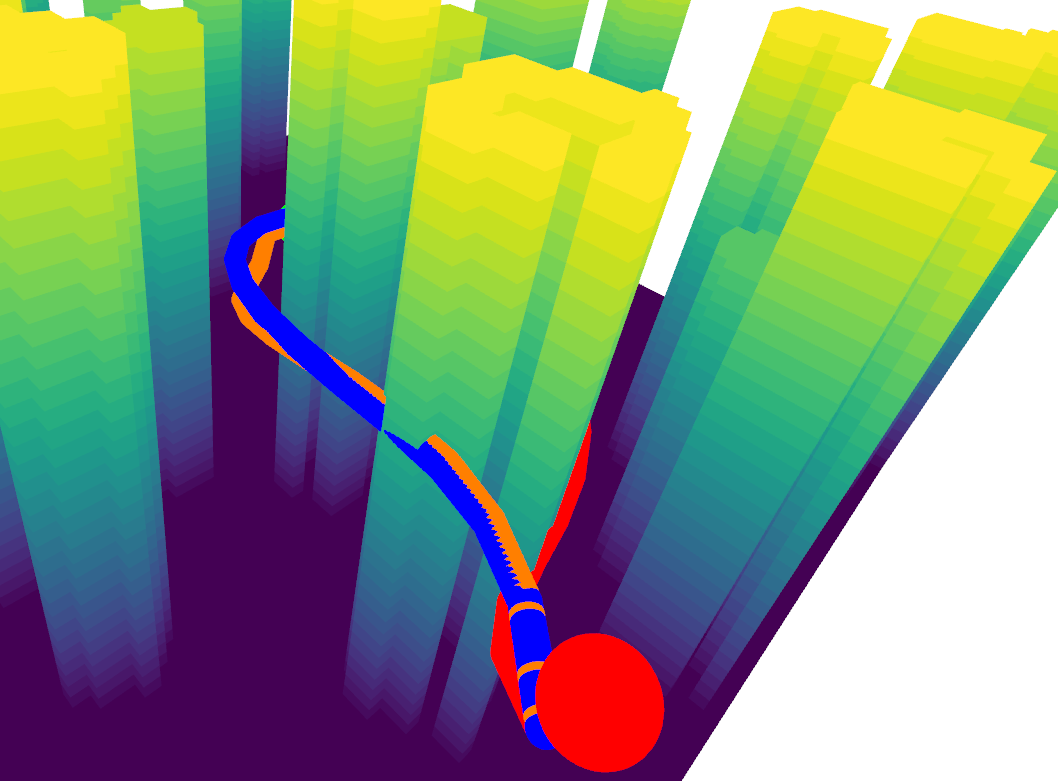}}
	\subfigure[]{\includegraphics[width=0.95in]{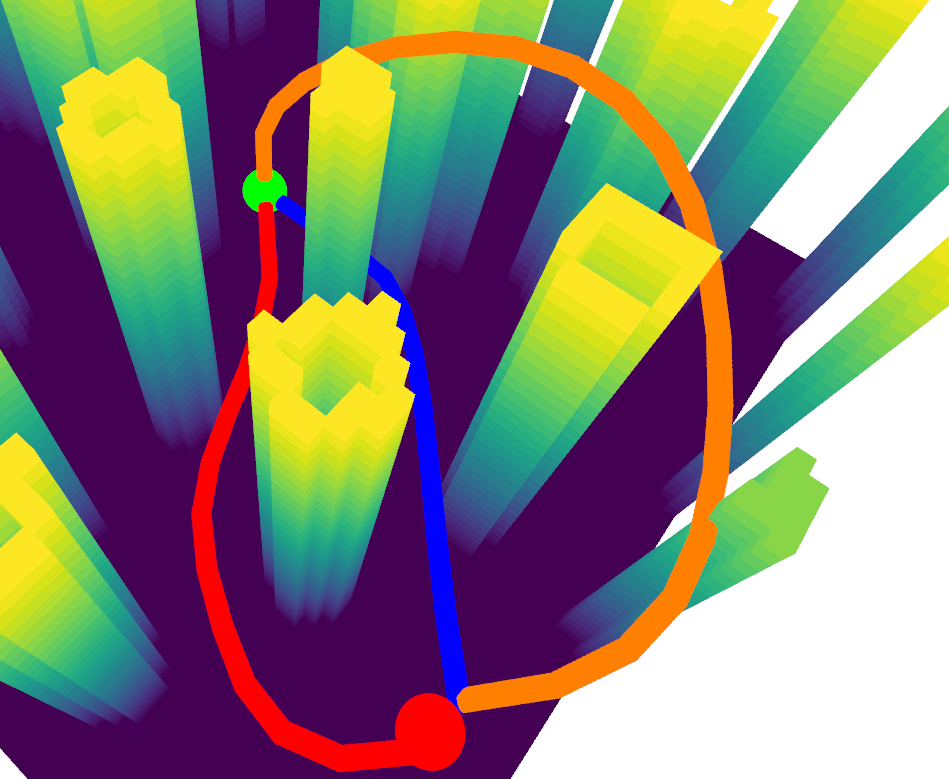}}
	\subfigure[]{\includegraphics[width=1.2in]{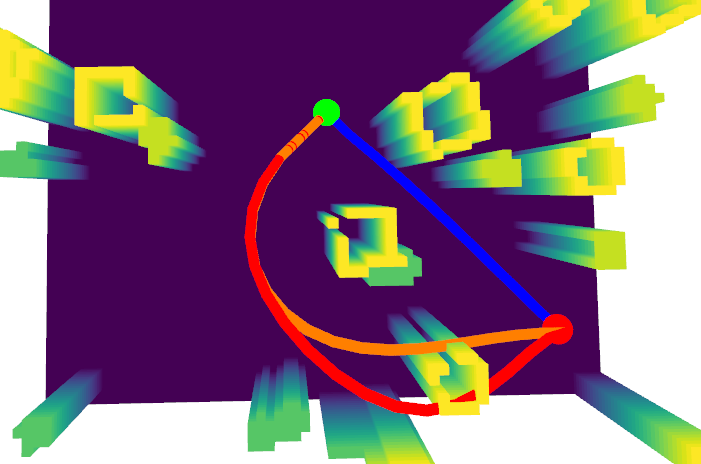}}
	\caption{Typical failure cases in pillar environments. Reconstruction errors in heavily occluded regions lead to inaccurate obstacle boundaries and eventually cause planning collisions.}
	\label{fig:failure_case}
\end{figure}

%

Typical failure cases are shown in Figures~\ref{fig:failure_case} and~\ref{fig:planning_compare}. Most failures arise from inaccurate reconstruction in heavily occluded or sparsely observed regions, where incomplete obstacle boundaries can lead to underestimated collision risks and unsafe trajectories. These results highlight the challenge of planning under partial observability and the importance of accurate local scene reconstruction.

\section{Conclusion}

This paper proposed a planning-oriented 3D scene completion framework that jointly 
learns TUDF and occupancy representations from partial observations. By enabling 
bidirectional interaction between continuous geometric estimation and discrete 
occupancy prediction, the proposed method achieves more complete and consistent 
scene reconstruction. Experiments in geometric and dense pillar environments 
demonstrate improved reconstruction and trajectory planning performance compared 
with single-task baselines. The learned TUDF can be directly integrated into 
trajectory optimization, providing a planning-friendly geometric representation 
under partial observability. Future work will explore uncertainty-aware scene 
completion to improve reasoning in highly unobserved regions.
	
\bibliographystyle{IEEEtran}
\bibliography{mylib}
\end{document}